\pdfoutput=1

\documentclass[11pt]{article}

\usepackage[]{acl}

\usepackage{times}
\usepackage{latexsym}

\usepackage[T1]{fontenc}

\usepackage[utf8]{inputenc}

\usepackage{microtype}

\usepackage{inconsolata}
\usepackage{booktabs}
\usepackage{graphicx}
\usepackage{tipa}
\usepackage{amssymb}
\usepackage{amsmath}
\usepackage{multirow}
\usepackage{subcaption}
\usepackage{pifont}

\title{A Likelihood Ratio Test of Genetic Relationship among Languages}

\author{V.S.D.S. Mahesh Akavarapu \and Arnab Bhattacharya \\
  Dept. of Computer Science and Engineering \\
  Indian Institute of Technology Kanpur \\
  \texttt{maheshak@cse.iitk.ac.in, arnabb@cse.iitk.ac.in}
}

\newcommand{\cmark}{\ding{51}}
\newcommand{\xmark}{\ding{55}}
\begin{document}

\maketitle

\begin{abstract}
Lexical resemblances among a group of languages indicate that the languages could be genetically related, i.e., they could have descended from a common ancestral language.  However, such resemblances can arise by chance and, hence, need not always imply an underlying genetic relationship. Many tests of significance based on permutation of wordlists and word similarity measures appeared in the past to determine the statistical significance of such relationships. We demonstrate that although existing tests may work well for bilateral comparisons, i.e., on pairs of languages, they are either infeasible by design or are prone to yield false positives when applied to groups of languages or language families. To this end, inspired by molecular phylogenetics, we propose a likelihood ratio test to determine if given languages are related based on the proportion of invariant character sites in the aligned wordlists applied during tree inference. Further, we evaluate some language families and show that the proposed test solves the problem of false positives. Finally, we demonstrate that the test supports the existence of macro language families such as Nostratic and Macro-Mayan.
\end{abstract}

\section{Introduction}
\label{sec:intro}
Languages that descend from a common ancestral language are termed to be \emph{genetically related}. The existence of lexical resemblances between the two languages is a preliminary indication that they could be related. Such resembling lexicons that truly have a common origin are called \emph{cognates}. For instance, Sanskrit \textit{n\={a}ma} and English \textit{name} are cognates that can be traced to Proto-Indo-European \textit{*h\textsubscript{3}n\'{o}mn}. However, such resemblances can also occur out of sheer chance. For instance, Persian \textit{bad} and \textit{behtar} accidentally resemble English \textit{bad} and \textit{better} respectively, but are not true cognates\footnote{Persian \textit{bad} is of uncertain origin while \textit{behtar} ultimately derives from PIE \textit{*h\textsubscript{1}w\'{e}sus}. On the other hand, English \textit{better} derives from PIE \textit{*b\textsuperscript{h}edr\'{o}s} and is cognate with Sanskrit \textit{bhadr\'{a}}}. Hence, it is necessary to show \emph{statistical significance} on any appropriate measure that captures the lexical relatedness before arguing for a genetic relationship among any group of languages or language families \cite{campbell2013historical}. 

Several significance tests appeared in the past to address this problem, with the majority of them based on permutation tests, starting from \citet{oswalt1970detection}. Given wordlists of a group of languages to be evaluated for a genetic relationship, these tests obtain the null distribution of a certain measure capturing similarity between word pairs by random permutations of the wordlists. Such tests either act \emph{bilaterally}, i.e., on a pair of languages or proto-languages, or \emph{multilaterally} on a group of languages. Among these, the multilateral comparison, which was made famous by \citet{greenberg1963languages, greenberg1971indo, greenberg1987language, greenberg2000indo} in traditional historical linguistics, has been a subject of much criticism \citep{poser2008language}. Hence, the preferred way of comparing two language families has been to compare their reconstructed proto-forms bilaterally. However, \citet{greenberg2005genetic} argues that genetic classification should precede proto-language reconstruction. Moreover, there is often a lack of agreement on reconstructed proto-forms both in terms of phonology and semantics which gives room for sufficient manipulation of wordlists that can in turn alter the results of significance tests \cite{kessler2015response}. Further, we demonstrate that multilateral permutation tests \citep{kessler2006multilateral, kessler-2007-word} yield false negatives even after incorporating complex word similarity metrics such as SCA and LexStat \citep{list2010sca, list-2012-lexstat}.

To overcome these issues, we turn to \emph{phylogenetic analysis} \cite{wiley2011phylogenetics} that is known to approximately capture the ancestral states and has been applied to phonological reconstruction tasks such as proto-language and cognate reflex prediction tasks \cite{jager2019computational, jager-2022-bayesian} with reasonably good results. Specifically, we propose a \emph{likelihood ratio test} (LRT) where we expect the difference in likelihoods of the best trees under null and alternate hypotheses to capture genetic relatedness. The null hypothesis assumes negligible proportion of invariant sites while the alternate hypothesis assumes significant proportion of invariant sites. Intuitively, related languages should have more positions where a character or a sound class is invariant than unrelated languages. Hence, we essentially capture the notion of relatedness as possessing a relatively high proportion of invariant sites. Further in this test, reconstructed proto-forms are not required and at the same time, the evolutionary tree structure is strictly imposed by design, unlike the multilateral model, thereby effectively circumventing the aforementioned methodological problems. Although inspired by similar tests from molecular phylogenetics, the test we propose is novel in the sense that the problem of testing common descent never arises in biology since monogenesis is accepted as a fact therein \citep{kessler2008mathematical}. We further evaluate the test on various language families and demonstrate that the test does not misclassify unrelated languages as related.

We finally show that the test supports the existence of the macro-families Nostratic \citep{bomhard1994nostratic} and Macro-Mayan \citep{campbell1997american}. While such an attempt to justify the existence of macro-families using bootstrap analysis of distance-based phylogeny is found in \citet{J_ger_2015}, expressing statistical significance in terms of likelihood ratio is preferred over bootstrap support values whose interpretation is debated in molecular phylogenetics \cite{anisimova2006approximate}.

Our contributions are summarized as follows.
\begin{itemize}
\item We have proposed a \emph{likelihood ratio test} to determine the \emph{genetic relatedness} of a group of languages based on \emph{invariant site proportions}.
\item We have demonstrated by applying various language sets that the test does not exhibit the problem of false positives nor requires reconstructed proto-forms, unlike the previously proposed tests.
\item We have found through the test some supporting evidence for the existence of macro-families namely Nostratic and Macro-Mayan
\end{itemize}

The rest of the paper is summarized as follows. Related work is discussed in \S\ref{sec:rel}. The methodology of the test is presented in \S\ref{sec:meth}. Evaluation details such as datasets and details of previous methods and variants are discussed in \S\ref{sec:exp}. The results are discussed in \S\ref{sec:res}. The application of the method on long-range comparisons is discussed in \S\ref{sec:mac}. The paper is concluded in \S\ref{sec:conc}.

\section{Related Work}
\label{sec:rel}


Permutation test for bilateral language relationship comparisons was introduced by \citet{oswalt1970detection}. The significance of sound correspondences by brute force probability calculation was proposed by \citet{ringe1992calculating, ringe1996mathematics}. This approach was however criticized for not being able to show significance for known related pairs of languages like Latin-English and also for accounting phonologically implausible sound correspondences \cite{kessler2001significance}. Multilateral permutation tests were proposed by \cite{kessler2006multilateral, kessler-2007-word}. Several applications of permutation tests exist such as \cite{turchin2010analyzing,kassian2015proto}.

Some notable likelihood ratio tests in molecular phylogenetics, mostly on topologies, include \cite{huelsenbeck1996likelihood,huelsenbeck1996likelihood1,goldman2000likelihood,anisimova2006approximate} where bootstrap analysis is argued to be not so optimal to establish statistical significance on phylogenies. Otherwise, support for macro-families through bootstrap analysis for distance-based trees is shown in \citet{J_ger_2015}. Comparisons of various methods of phylogenetic reconstruction such as distance-based and binary-character-based are given by \citet{J_ger_2018}. Sound-class character-based phylogenetic analysis is found in \cite{jager2019computational,jager-2022-bayesian}. Usually, Bayesian phylogenetic inference on binary cognate encodings gives good results \cite{rama-etal-2018-automatic, rama-list-2019-automated}.

Although the likelihood ratio metric is common for both past and present-day language models, the utility of this test using invariant sites outside computational historical linguistics is unknown.

\begin{figure*}[t]
    \centering
    \includegraphics[width=\textwidth]{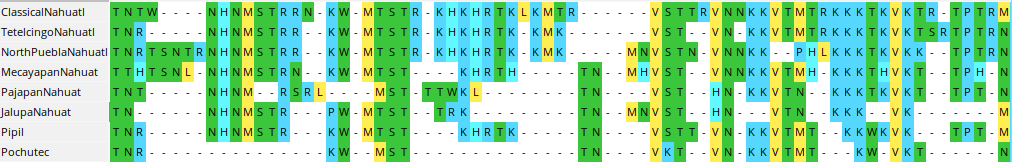}
    \caption{A section of character matrix for Uto-Aztecan family consisting of concatenated Multiple Sequence Alignments (MSAs) of consonant classes, one from each concept}
    \label{fig:charmat}
\end{figure*}

\section{Methodology}
\label{sec:meth}

The key concept revolves around the idea that any hypothesis, in this case, a hypothesis on a phylogeny, is preferred over a competing null hypothesis if it is significantly more likely, i.e., has a higher likelihood than the latter. Given the wordlist data encoded as an aligned character matrix, related languages are expected to have a higher number of \emph{invariant} columns. Thus, our null hypothesis consists of a phylogeny with a small proportion  (fixed at 1\%) of invariant sites, whereas the alternative hypothesis consists of a phylogeny with a larger but reasonable proportion (fixed at 6\%) of invariant sites. The observed difference in their likelihood of real data is compared with that of data simulated from the null hypothesis through parametric bootstrapping and, accordingly, one of the hypotheses is rejected. The steps are elaborated next.

\subsection{Character Matrix}
\label{subsec:charmat}

The wordlists of a given group of languages, as mentioned previously, are encoded in the form of a \emph{character matrix}. It consists of concatenated aligned words per concept, i.e., meaning. Thus, each row represents a language or \emph{taxon}, and each column, also referred to as \emph{site} in this paper, consists of phoneme classes, e.g., Dolgopolsky classes. Formally, let the input language set be $\{L_1, \ldots, L_m\}$, whose genetic relatedness is to be verified statistically. Let there be $n$ concepts $C_1, \ldots, C_n$ in the wordlists. Each language $L_i$ should have for each concept $C_j$ a single word, say $w_{ij}$. If a language has multiple words for a single semantic slot, only the one with fundamental or core meaning is retained, following the recipe by \citet{kessler2001significance}. For instance, if the meaning `dull' has words \textit{dull} and \textit{unsharp}, \textit{dull} is of core or fundamental meaning. Another example would be for the meaning `belly', Latin \textit{venter} is more fundamental than \textit{abd\={o}men}. If it so happens that it still remains unresolved after this step, a single word is randomly picked up. In case a language has no word for a semantic slot, it is represented as a gap `--'. For each concept $C_j$ and alphabet set $\mathbb{A}$, let $W^{j} \in \mathbb{A}^{m \times  l_{j}}$ represent a multiple sequence alignment (MSA) of words where $l_j$ is the length or the number of phonemes with vowels removed\footnote{Since the root form CVC is universal, including vowels results in spurious relationships. Further, languages of Caucasus like Georgian are rich in consonant clusters and, as a result, comparing them to others becomes difficult when vowels are considered.} in each word. The final character matrix $X \in \mathbb{A}^{m \times N}$ is concatenation of $W^{j}$, i.e., $[ W^{1} \ldots W^{n} ]$ across columns and  $N = \sum_{j=1}^{n}l_j$. 

\begin{table}[t]
    \centering
    \resizebox{0.65\columnwidth}{!}{
    \begin{tabular}{l|ccccc}
         \toprule
         {Greek\_Anc} & K & R & - & S \\
         {Latin} & K & R & N & - & - \\
         {English} &  H & R & N & - & - \\
         {Sanskrit} &  S & R & N & K & - \\
         \bottomrule
    \end{tabular}
    }
    \caption{Example of a Multiple Sequence Alignment (MSA) of consonant classes for a single concept `horn'.}
    \label{tab:msa}
\end{table}

For example, consider a cognate set meaning `horn' from a few Indo-European languages namely, Ancient Greek \textit{keras}, Latin \textit{cornu}, English \textit{horn}, and Sanskrit \textit{\'{s}\d{r}\.{n}ga}. The resultant character matrix for this single meaning is a multiple sequence alignment with vowels removed and consonants encoded as Dolgopolsky classes as illustrated in Table~\ref{tab:msa}. The final character matrix is the concatenation of such matrices across all the concepts. For an illustration of a final character matrix, see Figure~\ref{fig:charmat}, which is generated by MEGA11 \citep{tamura2021mega11}. In general, multiple sequence alignment is a fundamental step in several state-of-the-art methods in computational historical linguistics \citep{akavarapu-bhattacharya-2023-cognate, akavarapu-bhattacharya-2024-automated}.

\subsection{Substitution Model}
\label{subsec:subst}

A \emph{substitution model} describes the evolution of a character at a site assuming a Markovian process. Various substitution models have been described for various alphabets such as nucleotides, amino acids, etc. In this paper, we assume the simplest possible model where substitution rates are assumed to be equal between all the pairs of distinct characters. The resultant model is known as the Jukes-Cantor model \citep{jukes1969evolution} in case of nucleotide substitutions and as Poisson \citep{bishop1987tetrapod} in case of amino-acid substitutions. Formally, let the number of characters in the alphabet $\mathbb{A}$ be $N$. An element $q_{ij}$ of the rate matrix $Q$, which denotes the rate at which character $i$ mutates to character $j$ is defined as follows:
\begin{align}
q_{ij} = \mu \cdot \pi_{i} \mbox{ , } i \neq j \text{ (equal rates)}
\end{align}
where $\pi_{i}$ denotes the frequency of character $i$ at the site and $\mu$ is the rate of mutation. The diagonal element should satisfy the normalization constraint:
\begin{align}
q_{ii} = -\sum_{j \neq i}q_{ij}
\end{align}

The probability of transition $i \rightarrow j$ in time $t$ is given by the matrix $ P(t) = \{p_{ij}\} = e^{Qt}$. Likelihood of an evolutionary tree with topology $T$ can be, thus, calculated from the substitution matrix where branch lengths $V$ would denote the time.

\subsection{Maximum Likelihood Tree (ML-tree)}
\label{subsec:mltree}

For any phylogenetic tree with topology $T$, branch lengths $V$, other parameters such as shape parameter of heterogeneous rate, the proportion of invariant sites denoted by $\Theta$, and with the observed data i.e., character matrix $X$, the \emph{likelihood} is defined as the product of likelihoods at each site as given by the following equation, assuming independence for simplicity:
\begin{align}
\mathcal{L}(T,V,\Theta|X) = \prod_{i=1}^NP(X_i | T, V, \Theta)
\end{align}

The site independence assumption also restricts the number of parameters. Given the limited amount of data, which is restricted to 100-200 wordlists, this is, thus, more suitable. Complex models such as bigram-based ones may be employed if sufficient data is available.

The parameters that maximize the likelihood, $\hat{T}, \hat{V}$, and $\hat{\Theta}$, define the \emph{maximum likelihood tree} which is usually obtained by heuristic search in the parameter space. Typically, a tree is initialized either randomly or by some heuristic means, and from there, the tree space is explored through tree modifying operations to get the ``best'' tree. For a given tree, likelihood is computed using the well-known Felsenstein's pruning algorithm from phylogenetics \citep{felsenstein1973maximum, felsenstein1981evolutionary}.

\subsection{Invariant Sites}
\label{subsec:pinvar}

\emph{Invariant sites} are those sites that are constant or evolve very slowly. These can be estimated through a maximum likelihood search along with other parameters. The proportion of invariant sites, $P_{inv}$ may be known beforehand or estimated. Given the invariant sites, the likelihood defined in \S\ref{subsec:mltree} is only the product of likelihoods across the variant sites.

Our observation is that estimated $P_{inv}$ is higher ($>$0.06) among related languages while lower ($\approx$0.01) among (possibly) unrelated languages. Based on this observation and preliminaries, we now describe the likelihood ratio test.

\subsection{Likelihood Ratio Test (LRT)}
\label{subsec:lrt}

Given a null hypothesis $H_0$ and a competing alternative hypothesis $H_a$, the latter is preferred if it is more likely than the former i.e., $\mathcal{L}_{H_a} > \mathcal{L}_{H_0}$. In our case, the hypotheses consist of respective phylogenetic tree parameters estimated for ML-trees, i.e., $H_0$ consists of $\hat{T}_0, \hat{V}_0, \hat{\Theta}_0$ and $H_a$ consists of $\hat{T}_a, \hat{V}_a, \hat{\Theta}_a$. The likelihood ratio test defines the following metric to decide whether to reject the null hypothesis:
\begin{align}
\label{eqn:lrt}
\delta = 2 \cdot \mbox{ln}\left(\frac{\mathcal{L}(\hat{T}_a,\hat{V}_a,\hat{\Theta}_a)}{\mathcal{L}(\hat{T}_0,\hat{V}_0,\hat{\Theta}_0)}\right)
\end{align}

The \emph{Likelihood Ratio Test} (LRT) metric $\delta$ was shown to asymptotically follow a chi-squared distribution when the null hypothesis is assumed with the degrees of freedom $p-q$, where $p$ and $q$ respectively are the numbers of free parameters in the alternate and the null hypotheses \citep{Wilks_1938}. However, it was argued that this may not hold in general for phylogenetic problems due to the discrete nature of tree topology (see \cite{huelsenbeck1996likelihood, huelsenbeck1996likelihood1, anisimova2006approximate} for relevant work). As a result, the distribution of $\delta$ is determined by a parametric bootstrapping method where it is measured on the data simulated by the parameters estimated assuming the null hypothesis $H_0$ to hold, i.e, using the parameters $\hat{T}_0$, $\hat{V}_0$ and $\hat{\Theta}_0$.

As mentioned in \S\ref{subsec:pinvar}, we propose LRT to test the relatedness of a group of languages using varying proportions of invariant sites. In other words words the null hypothesis $H_0$ consists of invariant site proportion $P^0_{inv}$ and alternate hypothesis $H_a$ consists of $P^a_{inv}$ where $P^0_{inv} < P^a_{inv}$ as per the observations discussed in \S\ref{subsec:pinvar}. 

The typical way of obtaining the distribution for $\delta$ under $H_0$ involves finding the parameters $\{\hat{T}_0,\hat{V}_0,\hat{\Theta}_0\}$ and $\{\hat{T}_a,\hat{V}_a,\hat{\Theta}_a\}$ for the best trees respectively under $H_0$ and $H_a$ along with observed $\delta$, say $\hat{\delta}$. Further, several, say $k$, bootstrap replicates are generated from the topology, branch lengths, and other parameters defined by $\{\hat{T}_0,\hat{V}_0,\hat{\Theta}_0\}$, i.e., assuming $H_0$. Next, the maximum likelihood search is run again on these replicates to obtain several samples for $\delta$, say $\{\delta_1, \ldots, \delta_k\}$. However, we found considerable variation in $\hat{\delta}$, since the maximum likelihood search is only a heuristic and is affected by initialization. As a result, we obtain several samples for $\hat{\delta}$, say $\{\hat{\delta}_1, \ldots, \hat{\delta}_k\}$ by running the search $k$ times and based on the null parameters, a single bootstrap replicate is generated for each search to consequently obtain $\{\delta_1, \ldots, \delta_k\}$ for corresponding $k$ searches. Finally the \emph{p-value} for $\mathbb{E}[\delta] < \mathbb{E}[\hat{\delta}]$ is obtained by one-sided paired t-test. If the p-value is less than a threshold (usually 0.05), we conclude that $H_a$ may hold or, in other words, there are at least $P^a_{inv}$ proportions of sites that are significantly invariant and, thus, the languages under consideration are likely to be related.

\section{Experimental Setup}
\label{sec:exp}

The section discusses the details of the experiments including datasets, baseline models, and implementation details.

\subsection{Datasets}
\label{subsec:data}

\begin{table}[t]
\centering
\resizebox{\columnwidth}{!}{
\begin{tabular}{lcrrr}
\hline
\textbf{Family} & \textbf{Abbrv.} & \textbf{Languages} & \textbf{Concepts} & \textbf{Words} \\ \hline
Afrasian & AfA & 21 & 39 & 770 \\
Dravidian & Drav & 4 & 183 & 716 \\
Indo-European & IE & 12 & 185 & 2209 \\
Kartvelian & Kart & 1 & 180 & 180 \\
Lolo-Burmese & LoBur & 15 & 39 & 565 \\
Mayan & May & 30 & 94 & 2667 \\
Mixe-Zoque & MZ & 10 & 94 & 905 \\
Mon-Khmer & MKh & 9 & 199 & 1701 \\
Mon-Khmer & MKh & 16 & 94 & 1332 \\
Munda & Mun & 4 & 199 & 759 \\
Uto-Aztecan & UAz & 9 & 94 & 803 \\ \hline
\end{tabular}
}
\caption{Language families considered in this study.}
\label{tab:data}
\end{table}

The data for evaluating the tests consists of wordlists from multiple language (sub-)families and their combinations. Combinations of related sub-families serve as positive examples while those of unrelated serve as negative examples. Evaluating the macro-families also consists of language groups whose relationship is only distantly suggested such as Nostratic \citep{bomhard1994nostratic}.

The details of data from each family are shown in Table~\ref{tab:data}. Out of these, Mon-Khmer and Munda (200 wordlists) are extracted from the Austro-Asiatic data from \citet{rama-etal-2018-automatic}. Data for Old languages of Nostratic comprising Indo-European, Dravidian, and Kartvelian are prepared by us from the Swadesh 200-wordlists available at Wiktionary\footnote{\url{https://en.wiktionary.org/wiki/Category:Swadesh_lists_by_language}}. Data for all the other families are obtained from \citet{rama-2018-similarity} which were, in turn, collected from various publicly available sources. The datasets are the same as those found in related tasks such as automated cognate detection and proto-language reconstruction.

\addtocounter{footnote}{-1}

In the Nostratic grouping, we considered the languages that are surviving or have surviving descendants and were attested by the 10th century CE. The motivation behind this choice is that older languages should be closer to the ancestral language and each other if at all there is any relationship. Several languages including literary Dravidian languages, Georgian, and Armenian are mostly conservative and deviate little from their old forms. The data is pre-processed by excluding motivated word forms including onomatopoeia, and nursery forms, listed in \citet{kessler2001significance}. Short forms, i.e., words consisting of single syllables are also excluded. Such cleaning is necessary to avoid the appearance of spurious relationships. In the case of Nostratic, we were also careful to exclude borrowings by tracing etymologies from Wiktionary\footnotemark. This step could not be extended to other language families due to a lack of readily available etymological information.

All the methods employed in this work, including both the proposed one and baseline ones described in \S{\ref{subsec:kesslertest}}, involve the construction of a phylogenetic tree. Hence, we also compare the methods on a tree construction task where we see how well the trees match the golden truth trees wherever available. The data for this task is taken from \citet{rama-etal-2018-automatic} as summarized in Table~\ref{tab:treedata}.

\begin{table}[t]
\centering
\resizebox{\columnwidth}{!}{
\begin{tabular}{lcrrr}
\hline
\textbf{Family} & \textbf{Abbrv.} & \multicolumn{1}{l}{\textbf{Languages}} & \multicolumn{1}{l}{\textbf{Concepts}} & \multicolumn{1}{l}{\textbf{Words}} \\ \hline
Austro-Asiatic & AA & 58 & 200 & 11001 \\
Austronesian & AN & 45 & 210 & 8309 \\
Indo-European & IE & 42 & 208 & 8478 \\
Pama-Nyungan & PN & 67 & 183 & 11503 \\
Sino-Tibetan & ST & 64 & 110 & 6762 \\ \hline
\end{tabular}
}
\caption{Language family datasets for tree construction.}
\label{tab:treedata}
\end{table}

\subsection{Multilateral Permutation Test}
\label{subsec:kesslertest}

As mentioned in \S\ref{sec:intro}, most previous methods compare languages bilaterally, i.e., a pair at a time. As a result, the only possible way to compare the language families in this approach is to compare their reconstructed proto-languages. However, proto-forms of a proto-language are not often universally agreed which leads to considerable allowance of manipulation that can affect the results \citep{kessler2015response}. An alternate solution to determine the significance of the relationship among multiple languages was proposed by \citet{kessler2006multilateral} and \citet{kessler-2007-word} who employ a permutation test based on multilateral comparison. This has been well received in historical linguistics \citep{ringe2013historical}.

The test is based on nearest-neighbour hierarchical clustering where at any point two closest clusters are lumped into one cluster. The basic distance measure, $\hat{d}(A,B)$, between any two clusters $A$ and $B$ is the average of distances between all possible pairs of languages in these clusters, i.e.,
\begin{align}
\hat{d}(A,B) = \frac{1}{|A|\cdot|B|}\sum_{a \in A}\sum_{b \in B}d(a,b)
\end{align}
where the distance $d(a,b)$ between any two languages $a$ and $b$ is the mean distance between the pairs of words over all concepts. Following the notations of \S\ref{subsec:charmat} where $w_{aj}$ and $w_{bj}$ are words in languages $a$ and $b$ respectively from concept $C_j$,
\begin{align}
d(a,b) = \frac{\sum_{C_j, w_{aj} \neq \emptyset, w_{bj} \neq \emptyset} d(w_{aj}, w_{bj})}{|\{C_j :  w_{aj} \neq \emptyset, w_{bj} \neq \emptyset \}|}
\end{align}

Taking an average over all languages essentially enforces multilateral comparison, i.e., multiple languages are being considered equally to compute the outcome. Further, the algorithm thus described is the same as UPGMA tree construction method \citep{Sokal1958ASM} where at any bifurcating node, a uniform rate of evolution is assumed across daughter clades. The final similarity metric $\hat{s}(A,B)$ is determined by the following statistic that is computed based on a random permutation of words across each column (taxon) which yields random distances $d(A,B)$:
\begin{align}
\label{eqn:mptsim}
\hat{s}(A,B) = \frac{\mathbb{E}[d(A,B)] - \hat{d}(A,B)}{\mathbb{E}[d(A,B)]}
\end{align}

The \emph{p-value} of two language clusters $A$ and $B$ is the frequency of the event $\hat{d}(A,B) \geq d(A,B)$ relative to the total number of random permutations.
Language clusters $A$ and $B$ are considered to be \emph{related} if the p-value is less than $0.05$. The given languages are termed \emph{related} if the final two clusters that are merged at the root are related \citep{kessler2006multilateral}. 

\citet{kessler-2007-word} ran this test using various word similarity metrics which almost give similar results. Among these metrics, we ran on P1-dolgo which is a binary metric that determines whether the consonant class of the word's initial consonant matches or not. Additionally, we employ the binary similarity measure introduced by \citet{turchin2010analyzing} to test the significance of the Altaic family where the first two consonants are considered. We further test continuous word distances introduced by \citet{list2010sca} (SCA) and \citet{list-2012-lexstat} (LexStat) that are based on sequence alignment techniques which were introduced in the context of automated cognate detection.

\begin{table*}[t]
\centering
\resizebox{\textwidth}{!}{
\begin{tabular}{l||cccccccc|ccc}
\hline
\textbf{Method} & \textbf{MKh} & \textbf{Mun} & \textbf{MKh-Mun}  & \textbf{IE} & \textbf{Drav} & \textbf{May} & \textbf{MZ} & \textbf{UAz} & \textbf{MKh-May} & \textbf{MKh-UAz} & \textbf{AfA-LoBur} \\ \hline\hline
\textbf{Related} & \cmark & \cmark & \cmark & \cmark & \cmark & \cmark & \cmark & \cmark & \xmark & \xmark & \xmark  \\ \hline
\textbf{P1-Dolgo} & \begin{tabular}[c]{@{}c@{}}0.123\\ (\textless{}0.001)\end{tabular} & \begin{tabular}[c]{@{}c@{}}0.243\\ (\textless{}0.001)\end{tabular} & \begin{tabular}[c]{@{}c@{}}0.080\\ (\textless{}0.001)\end{tabular}  & \begin{tabular}[c]{@{}c@{}}0.071\\ (\textless{}0.001)\end{tabular} & \begin{tabular}[c]{@{}c@{}}0.440 \\ (\textless{}0.001)\end{tabular} & \begin{tabular}[c]{@{}c@{}}0.228 \\ (\textless{}0.001)\end{tabular} & \begin{tabular}[c]{@{}c@{}}0.412 \\ (\textless{}0.001)\end{tabular} & \begin{tabular}[c]{@{}c@{}}0.572 \\ (\textless{}0.001)\end{tabular} & {\color[HTML]{FF0000} \begin{tabular}[c]{@{}c@{}}0.007\\ (\textless{}0.001)\end{tabular}} & \begin{tabular}[c]{@{}c@{}}0.005\\ (0.063)\end{tabular} & {\color[HTML]{FF0000} \begin{tabular}[c]{@{}c@{}}0.017\\ (\textless{}0.001)\end{tabular}} \\ \hline
\textbf{Turchin} & \begin{tabular}[c]{@{}c@{}}0.019 \\ (\textless{}0.001)\end{tabular} & \begin{tabular}[c]{@{}c@{}}0.124 \\ (\textless{}0.001)\end{tabular} & \begin{tabular}[c]{@{}c@{}}0.019 \\ (\textless{}0.001)\end{tabular}  & \begin{tabular}[c]{@{}c@{}}0.028 \\ (\textless{}0.001)\end{tabular} & \begin{tabular}[c]{@{}c@{}}0.292 \\ (\textless{}0.001)\end{tabular} & \begin{tabular}[c]{@{}c@{}}0.126 \\ (\textless{}0.001)\end{tabular} & \begin{tabular}[c]{@{}c@{}}0.256 \\ (\textless{}0.001)\end{tabular} & \begin{tabular}[c]{@{}c@{}}0.402 \\ (\textless{}0.001)\end{tabular} & {\color[HTML]{FF0000} \begin{tabular}[c]{@{}c@{}}0.003 \\ (\textless{}0.001)\end{tabular}} & {\color[HTML]{FF0000} \begin{tabular}[c]{@{}c@{}}0.003 \\ (0.005)\end{tabular}} & {\color[HTML]{FF0000} \begin{tabular}[c]{@{}c@{}}0.004 \\ (\textless{}0.001)\end{tabular}} \\ \hline
\textbf{LexStat} & \begin{tabular}[c]{@{}c@{}}0.065 \\ (\textless{}0.01)\end{tabular} & \begin{tabular}[c]{@{}c@{}}0.138 \\ (\textless{}0.01)\end{tabular} & \begin{tabular}[c]{@{}c@{}}0.048 \\ (\textless{}0.01)\end{tabular}  & \begin{tabular}[c]{@{}c@{}}0.036 \\ (\textless{}0.01)\end{tabular} & \begin{tabular}[c]{@{}c@{}}0.197 \\ (\textless{}0.01)\end{tabular} & \begin{tabular}[c]{@{}c@{}}0.129 \\ (\textless{}0.01)\end{tabular} & \begin{tabular}[c]{@{}c@{}}0.244 \\ (\textless{}0.01)\end{tabular} & \begin{tabular}[c]{@{}c@{}}0.306 \\ (\textless{}0.01)\end{tabular} & {\color[HTML]{FF0000} \begin{tabular}[c]{@{}c@{}}0.028 \\ (\textless{}0.01)\end{tabular}} & {\color[HTML]{FF0000} \begin{tabular}[c]{@{}c@{}}0.018 \\ (\textless{}0.01)\end{tabular}} & {\color[HTML]{FF0000} \begin{tabular}[c]{@{}c@{}}0.033 \\ (\textless{}0.01)\end{tabular}} \\ \hline
\textbf{SCA} & \begin{tabular}[c]{@{}c@{}}0.087 \\ (\textless{}0.01)\end{tabular} & \begin{tabular}[c]{@{}c@{}}0.187 \\ (\textless{}0.01)\end{tabular} & \begin{tabular}[c]{@{}c@{}}0.074 \\ (\textless{}0.01)\end{tabular}  & \begin{tabular}[c]{@{}c@{}}0.056 \\ (\textless{}0.01)\end{tabular} & \begin{tabular}[c]{@{}c@{}}0.296 \\ (\textless{}0.01)\end{tabular} & \begin{tabular}[c]{@{}c@{}}0.177 \\ (\textless{}0.01)\end{tabular} & \begin{tabular}[c]{@{}c@{}}0.304 \\ (\textless{}0.01)\end{tabular} & \begin{tabular}[c]{@{}c@{}}0.400 \\ (\textless{}0.01)\end{tabular} & {\color[HTML]{FF0000} \begin{tabular}[c]{@{}c@{}}0.015 \\ (\textless{}0.01)\end{tabular}} & {\color[HTML]{FF0000} \begin{tabular}[c]{@{}c@{}}0.006 \\ (\textless{}0.01)\end{tabular}} & {\color[HTML]{FF0000} \begin{tabular}[c]{@{}c@{}}0.031 \\ (\textless{}0.01)\end{tabular}} \\
\hline\hline
\textbf{LRT} & \begin{tabular}[c]{@{}c@{}}9.205 \\ (\textless{}0.001)\end{tabular} & \begin{tabular}[c]{@{}c@{}}1.58 \\ (\textless{}0.001)\end{tabular} & \begin{tabular}[c]{@{}c@{}}14.18 \\ (\textless{}0.001)\end{tabular}  & \begin{tabular}[c]{@{}c@{}}26.154 \\ (\textless{}0.001)\end{tabular} & \begin{tabular}[c]{@{}c@{}}1.78 \\ (\textless{}0.001)\end{tabular} & \begin{tabular}[c]{@{}c@{}}68.212 \\ (\textless{}0.001)\end{tabular} & \begin{tabular}[c]{@{}c@{}}7.192 \\ (\textless{}0.001)\end{tabular} & \begin{tabular}[c]{@{}c@{}}10.448 \\ (\textless{}0.001)\end{tabular} & \begin{tabular}[c]{@{}c@{}}-14.359 \\ (0.280)\end{tabular} & \begin{tabular}[c]{@{}c@{}}-12.188 \\ (0.065)\end{tabular} & \begin{tabular}[c]{@{}c@{}}-10.768 \\ (0.979)\end{tabular} \\ \hline\hline
\end{tabular}
}
\caption{Significance testing on various existent and non-existent families. The values indicate the similarity measure $\hat{s}$ in the case of permutation tests and in the case of LRT they indicate the mean of statistic $\hat{\delta}$. Values in parentheses indicate p-value. False positives are marked in {\color[HTML]{FF0000}red}.}
\label{tab:restest}
\end{table*}

\subsection{Implementation}
\label{subsec:impl}

We mapped the consonant classes to the protein alphabet since phylogenetic software expects input as either nucleotide or amino acid sequences. Moreover, most of the amino acid letters and Dolgopolsky classes are identical. In this regard, there is only one exception, namely, `J' which is absent in the former but present in the latter and is, hence, simply replaced with `I', which is in turn absent in Dolgopolsky classes. The multiple sequence alignments are obtained from CLUSTALW2 \citep{larkin2007clustal} while the best trees and their corresponding likelihoods were computed using IQ-TREE \citep{nguyen2015iq}. As described in \S\ref{subsec:pinvar} and \S\ref{subsec:lrt}, the proportions of invariant sites $P_{inv}^0$ and $P_{inv}^a$ are set to 0.01 and 0.06 respectively for null ($H_0$) and alternate ($H_a$) hypotheses. The parametric bootstrap replicates are generated using AliSim \citep{ly2022alisim}, an extension of IQ-TREE. To replicate as closely as possible, gaps present in the original character matrices are retained in the replicates. We calculate the p-value based on a sample size of $k = 15$. The outcomes are observed to be stable beyond this size.
The word similarity metrics used in the baseline models are computed by using Lingpy \citep{list2021lingpy}. For the phylogenetic tree construction task, MEGA11 \citep{tamura2021mega11} was used to deduce the maximim likelihood tree (ML-tree) with the aforementioned model with an additional gamma rate heterogeneity parameter with two distinct rates whose shape is estimated. We name this method \emph{ML-P+I+G2}.

The \emph{generalized quartet distances} (GQD) \citep{pompei2011accuracy} between the predicted and the gold trees are computed from quartet distances obtained using qdist \citep{mailund2004qdist}. The \emph{quartet distance} between two trees measures the number of four-leaf-subsets that have dissimilar topologies. Unlike biological phylogenetic trees, language trees are often multifurcated. Hence, GQD excludes penalties over the order of bifurcations. The code and relevant data have been made publicly available\footnote{\url{https://github.com/mahesh-ak/PhyloVal}}. Further implementation details can be found in \texttt{README.md} therein.

\section{Results}
\label{sec:res}

The primary results of the paper are tabulated in Table \ref{tab:restest}, where the results of LRT (last row) are compared with those of the multilateral permutation tests. Except for LRT, the column `Method' indicates the distance metric employed in the permutation test. The row `Related' indicates the current consensus about the relatedness of the language families. For the permutation test, the values indicate the similarity metric $\hat{s}$ defined in Eq.~\eqref{eqn:mptsim}, as measured at the root. On the other hand, for LRT the values indicate the mean of observed $\hat{\delta}$ (see \S{\ref{subsec:lrt}}). The p-values are indicated in parentheses. The standard threshold of 0.05 is assumed for p-values. Please refer to Table~\ref{tab:data} and Table~\ref{tab:treedata} for abbreviations of various language families.

One can observe that false positives, indicated in red, are absent for LRT, in contrast with multilateral permutation tests which exhibit false positives in all cases (except P1-Dolgo for MKh-UAz). However, we note that the similarity scores of the Turchin measure are consistently small ($<0.005$) for negatives irrespective of the significance implied by the p-value. Hence, it may be noted that Turchin could be a good measure for permutation tests when similarity scores are taken into consideration.

Further, one can observe from Table \ref{tab:restest} that mean $\hat{\delta}$ values are small for valid families such as Mun and Drav. This has to do with the fact that the data for these families consists of a lower number of taxa (see Table~\ref{tab:data}). Hence, althought the $\hat{\delta}$ measure need not imply strength, its sign implies which hypothesis is to be preferred, i.e., the one with a larger proportion of invariant sites in case of a positive value and the one with a smaller proportion of invariant sites in case of a negative value.

\subsection{Tree Construction}
\label{subsec:tree}

\begin{table}[t]
\centering
\resizebox{\columnwidth}{!}{
\begin{tabular}{l|ccccc|c}
\hline
\textbf{Method} & \textbf{AA} & \textbf{AN} & \textbf{IE} & \textbf{PN} & \textbf{ST} & \textbf{Avg} \\ \hline
\textbf{P1-Dolgo} & 0.060 & 0.208 & 0.033 & 0.175 & 0.188 & 0.133 \\
\textbf{Turchin} & 0.069 & 0.195 & 0.058 & 0.175 & 0.275 & 0.154 \\
\textbf{LexStat} & 0.051 & 0.178 & \textbf{0.020} & 0.164 & 0.096 & 0.102 \\
\textbf{SCA} & 0.049 & 0.119 & 0.025 & 0.166 & \textbf{0.087} & 0.089 \\
\hline
\textbf{ML-P+I+G2} & \textbf{0.026} & \textbf{0.065} & 0.033 & \textbf{0.145} & 0.125 & \textbf{0.079} \\ \hline
\end{tabular}
}
\caption{Comparison of the methods on phylogenetic tree construction task provided as GQD scores. The best results are in \textbf{bold}.}
\label{tab:treeres}
\end{table}

As mentioned in \S\ref{subsec:data}, both the methods output a tree, and, therefore, the methods have been evaluated on the tree construction task. The purpose of this task is to ensure that the proposed methods have indeed a good sense of phylogenetic inference and are, hence, appropriate to carry out significance tests over phylogenies. The results are tabulated in Table \ref{tab:treeres}. By comparing with the mean scores of state-of-the-art language phylogeny inference methods on this data, ML-P+I+G2 (0.079) is a few steps behind Bayesian inferred tree (0.066) \citep{rama-etal-2018-automatic} and maximum a posteriori tree (0.051) \citep{rama-list-2019-automated}. Hence, it can be concluded that consonant-class-based character matrix encoding is almost as good as cognate-based binary character matrix encoding while probabilistic methods based on character matrices are superior to distance-based methods for this task. Among the distance-based approaches, one with the SCA metric performs best. A similar situation was observed in \citet{rama-etal-2018-automatic} and \citet{rama-list-2019-automated} where SCA-based cognates yield the best performance. However, it should be noted that SCA and LexStat-based measures yield false positives on significance testing (Table~\ref{tab:restest}) despite their performance on this task. 

\section{Evaluation of Macro Families}
\label{sec:mac}

\begin{table}[t]
\centering
\resizebox{\columnwidth}{!}{
\begin{tabular}{l|ccccc}
\hline
\textbf{Method} & \textbf{Drav-IE} & \textbf{Drav-IE-Kart} & \textbf{May-MZ} & \textbf{May-UAz} & \textbf{May-MZ-UAz} \\ \hline\hline
\textbf{P1-Dolgo} & \begin{tabular}[c]{@{}c@{}}0.046 \\ (\textless{}0.001)\end{tabular} & \begin{tabular}[c]{@{}c@{}}0.038 \\ (\textless{}0.001)\end{tabular} & \begin{tabular}[c]{@{}c@{}}0.033 \\ (\textless{}0.001)\end{tabular} & \begin{tabular}[c]{@{}c@{}}0.046 \\ (\textless{}0.001)\end{tabular} & \begin{tabular}[c]{@{}c@{}}0.036 \\ (\textless{}0.001)\end{tabular} \\ \hline
\textbf{Turchin} & \begin{tabular}[c]{@{}c@{}}0.017 \\ (\textless{}0.001)\end{tabular} & \begin{tabular}[c]{@{}c@{}}0.002 \\ (0.197)\end{tabular} & \begin{tabular}[c]{@{}c@{}}0.012 \\ (\textless{}0.001)\end{tabular} & \begin{tabular}[c]{@{}c@{}}0.012 \\ (\textless{}0.001)\end{tabular} & \begin{tabular}[c]{@{}c@{}}0.008 \\ (\textless{}0.001)\end{tabular} \\ \hline
\textbf{LexStat} & \begin{tabular}[c]{@{}c@{}}0.024 \\ (\textless{}0.01)\end{tabular} & \begin{tabular}[c]{@{}c@{}}0.014 \\ (\textless{}0.01)\end{tabular} & \begin{tabular}[c]{@{}c@{}}0.033 \\ (\textless{}0.01)\end{tabular} & \begin{tabular}[c]{@{}c@{}}0.027 \\ (\textless{}0.01)\end{tabular} & \begin{tabular}[c]{@{}c@{}}0.024 \\ (\textless{}0.01)\end{tabular} \\ \hline
\textbf{SCA} & \begin{tabular}[c]{@{}c@{}}0.024 \\ (\textless{}0.01)\end{tabular} & \begin{tabular}[c]{@{}c@{}}0.007 \\ (0.01)\end{tabular} & \begin{tabular}[c]{@{}c@{}}0.019 \\ (\textless{}0.01)\end{tabular} & \begin{tabular}[c]{@{}c@{}}0.024 \\ (\textless{}0.01)\end{tabular} & \begin{tabular}[c]{@{}c@{}}0.015 \\ (\textless{}0.01)\end{tabular} \\
\hline\hline
\textbf{LRT} & \begin{tabular}[c]{@{}c@{}}24.882 \\ (\textless{}0.001)\end{tabular} & \begin{tabular}[c]{@{}c@{}}0.316 \\ (\textless{}0.001)\end{tabular} & \begin{tabular}[c]{@{}c@{}}20.988 \\ (\textless{}0.001)\end{tabular} & \begin{tabular}[c]{@{}c@{}}-1.035 \\ (\textless{}0.001)\end{tabular} & \begin{tabular}[c]{@{}c@{}}-9.819 \\ (\textless{}0.001)\end{tabular} \\ \hline\hline
\end{tabular}
}
\caption{Results of evaluation of macro families. Parentheses contain p-values.}
\label{tab:mac}
\end{table}

We apply the tests on groupings of a few families from proposed macro families, namely Nostratic, Macro-Mayan, and Amerind. Under Nostratic, we test for groupings Dravidian-Indo-European (\emph{Drav-IE}) and Dravidian-Indo-European-Kartvelian (\emph{Drav-IE-Kart}) while we test Mayan-Mixe-Zoque (\emph{May-MZ}) under Macro-Mayan and Mayan-Uto-Aztecan (\emph{May-UAz}), Mayan-Mixe-Zoque-Uto-Aztecan (\emph{May-MZ-UAz}) under Amerind. The results are tabulated in Table~\ref{tab:mac}. While going by the p-values, the LRT test seems to support all of the mentioned families. However, the mean LRT statistic $\hat{\delta}$ is weak (negative or close to $0$) for Drav-IE-Kart (Nostratic) and May-UAz, May-MZ-UAz (Amerind). In other words, by looking at Eq.~\eqref{eqn:lrt}, the alternate hypothesis $H_a$, i.e., having higher invariant sites is not preferred. Thus, it may be concluded that LRT is a highly sensitive test since the mere addition of a single language (Georgian) to a strongly supported group of 16 languages (Drav-IE) alters the outcome drastically. This is a desirable property since the presence of even a single anomaly, an unrelated language in this case, can be detected. Note that other combinations in Nostratic such as Drav-Kart or IE-Kart are much weaker and not well supported by the permutation test itself, which is elaborated as follows.

\subsection{Analysis of Permutation tests on Nostratic}
\label{subsec:biltrl}

\begin{figure*}[t]
    \centering
    \begin{subfigure}[t]{0.24\textwidth}
        \centering
        \includegraphics[width=\textwidth]{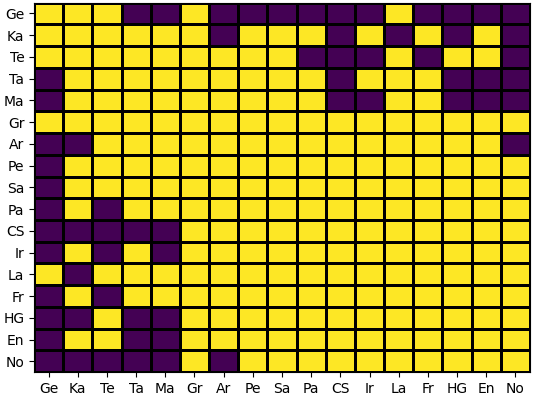}
        \caption{P1-Dolgo}
    \end{subfigure}%
    ~ 
   \begin{subfigure}[t]{0.24\textwidth}
        \centering
        \includegraphics[width=\textwidth]{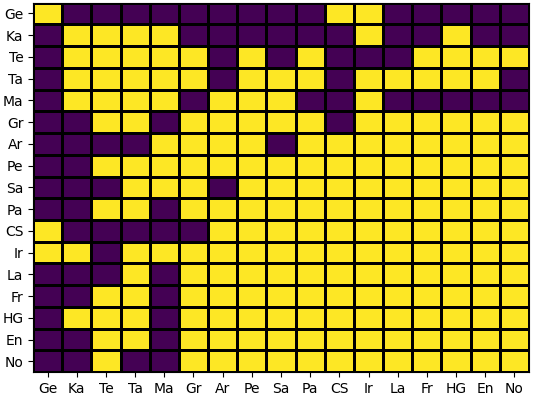}
        \caption{Turchin}
    \end{subfigure}%
    ~ 
    \begin{subfigure}[t]{0.24\textwidth}
        \centering
        \includegraphics[width=\textwidth]{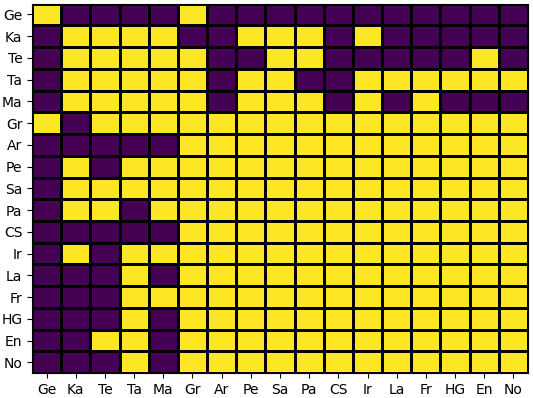}
        \caption{SCA}
    \end{subfigure}%
    ~ 
    \begin{subfigure}[t]{0.24\textwidth}
        \centering
        \includegraphics[width=\textwidth]{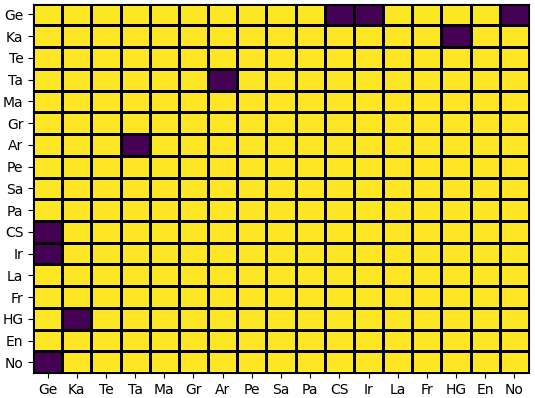}
        \caption{LexStat}
    \end{subfigure}%
    \caption{Bilateral (pairwise) significance among the languages of Nostratic grouping. The yellow shade implies that the relationship is statistically significant ($p < 0.05$), while the purple shade implies otherwise.}
    \label{fig:biltrl}
\end{figure*}

Bilateral significances on Nostratic grouping Drav-IE-Kart for various distance metrics are reported in Figure~\ref{fig:biltrl}, where the pairwise relationships based on p-value (with threshold 0.05) are color-coded. The computation follows the same steps as defined in \S\ref{subsec:kesslertest} except that distances and similarities are calculated over pairs of languages instead of language clusters. This indeed forms the first iteration of a complete multilateral test. 

The languages are abbreviated in Fig.~\ref{fig:biltrl} as follows:  Old Georgian (Ge), Old Kannada (Ka), Old Telugu (Te), Old Tamil (Ta), Old Malayalam (Ma), Ancient Greek (Gr), Old Armenian (Ar), Middle Persian (Pe), Sanskrit (Sa), Pali (Pa), Old Church Slavonic (CS), Old Irish (Ir), Latin (La), Old French (Fr), Old High German (HG), Old English (En) and Old Norse (No). 

It is visible that for each metric, languages of the same family (IE and Drav) are almost always related pairwise. Secondly, many pairs from Drav-IE appear related. However, except for LexStat, Georgian shows to be related to at most two languages from the Drav-IE grouping. Yet, in the permutation tests for these metrics, except for Turchin (Table \ref{tab:mac}), Drav-IE-Kart appears significantly related with sometimes even good similarity scores (in the case of P1-Dolgo). All that can be concluded here is that, except for the LexStat metric, permutation tests are very sensitive to pairwise language comparisons and may not yield false positives. However, if Drav-IE-Kart is to be considered a valid grouping, these tests may be said to yield false negatives.

\subsection{Analysis of ML-trees of Nostratic}
\label{subsec:nosmltree}

\begin{figure*}[t]
    \centering
    \begin{subfigure}[t]{0.32\textwidth}
        \centering
        \includegraphics[width=\textwidth]{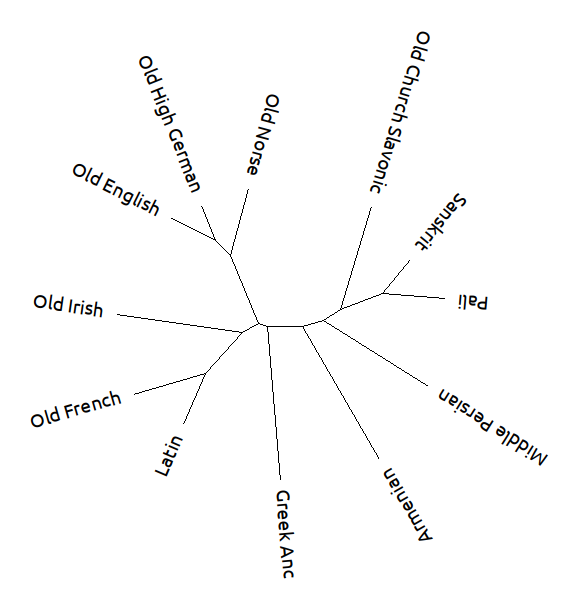}
        \caption{IE}
    \end{subfigure}%
    ~ 
    \begin{subfigure}[t]{0.30\textwidth}
        \centering
        \includegraphics[width=\textwidth]{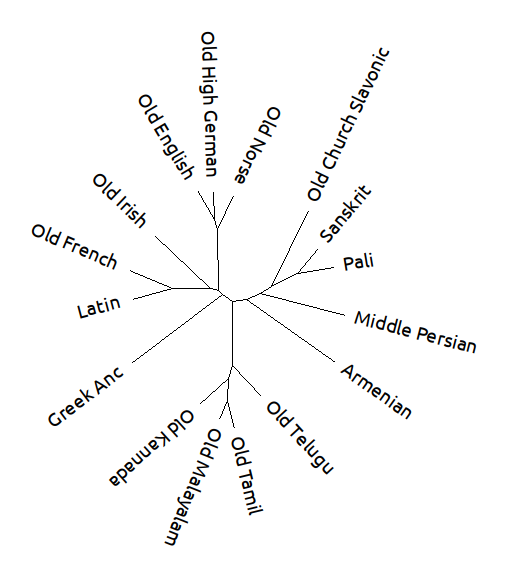}
        \caption{Drav-IE}
    \end{subfigure}
    ~ 
    \begin{subfigure}[t]{0.30\textwidth}
        \centering
        \includegraphics[width=\textwidth]{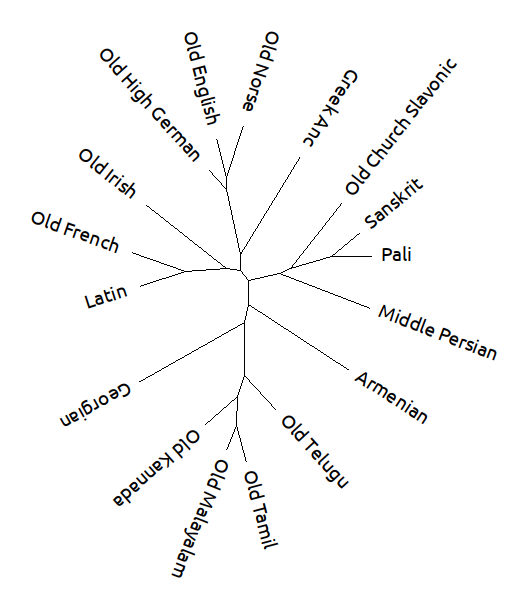}
        \caption{Drav-IE-Kart}
    \end{subfigure}
    \caption{Comparison of unrooted ML-trees on various groupings of Nostratic language families}
    \vspace*{-2mm}
    \label{fig:mltree}
\end{figure*}

Unrooted maximum likelihood trees (ML-trees) are drawn in Figure~\ref{fig:mltree} on various sub-groupings of Nostratic using MEGA11 assuming the Poisson+I model. For the IE tree (Figure~\ref{fig:mltree}(a)), the sub-families, except for the position of Old Church Slavonic, are highly faithful reflecting the existing notions. For instance, the topology of the Germanic family, i.e., (Old Norse, (Old English, Old High German)) contains the valid West-Germanic branch (Old English, Old High German). Similarly, the Italo-Celtic group (Old Irish, (Latin, Old French)) is visible. Also, one can distinguish a clear boundary between Western and Eastern IE languages reflecting the geographical distribution. However, the position of Old Church Slavonic intruded into Indo-Iranian appears problematic.

Further, the addition of the Dravidian family in Drav-IE does not alter the IE topology (Figure~\ref{fig:mltree}(b)). It is intriguing to note the western inclination of Dravidian given its eastern geographical location in the present day. However, this is in line with the observation of \citet{caldwell1875comparative}, the founder of comparative Dravidian linguistics himself. Finally, the addition of Georgian invalidates the West-Germanic branch as well as pushes Old Greek problematically into the Western group (Figure~\ref{fig:mltree}(c)). However, much of the topology is undisturbed and one can also notice how the languages/families that are located south of the Caucasus namely, Armenian, Georgian, and Dravidian are grouped. Overall, it may be concluded that the addition of unrelated or weakly related languages can alter the actual topology.

Similar analyses in case of Macro-Mayan and Amerind families are provided in Appendix \ref{sec:appendix} where one can observe similar perturbations in topology (see Fig. \ref{fig:mmu_mltree}) of one family (Mayan) in presence of others (Mixe-Zoque and Uto-Aztecan).
    
\section{Conclusions}
\label{sec:conc}

In this paper, we have presented a likelihood ratio test based on the proportions of invariant sites to determine the genetic relatedness of a group of languages. Our proposed test does not yield false positives, which is in contrast with previous permutation-based tests that proved to be good only for pairwise language comparisons and not for validating a language group. By applying this test, we have found strong supporting evidence for macro-families such as Dravidian-Indo-European, Macro-Mayan (for Mayan-Mixe-Zoque, and weak evidence for Nostratic (Dravidian-Indo-European-Kartvelian) and Amerind (for Mayan-Uto-Aztecan). Through secondary analyses, we have also shown that probabilistic-based methods are superior to distance-based ones based on tree construction and the correlation of topologies with geography. In this work we did not touch upon semantic shifts, i.e., words changing meaning over time; for example, the word \textit{quick} initially meant `lively'. While considering semantic shifts may provide room for data manipulation favoring any particular hypothesis, few semantic slots such as `bark'-`skin' are often found to have common words. In such cases, the slots may be merged into one as suggested by \citet{kessler2001significance}.

In summary, before constructing phylogenies of a group of languages, the relatedness of the group should be established through a significance test such as the one we have presented. Otherwise, the phylogenic grouping would not only be questionable but may also alter the topology of a related sub-group.


\section*{Limitations}

The values of $P^0_{inv}$ and $P^a_{inv}$ (\S\ref{subsec:lrt}) are roughly decided based on the estimated ones from two examples, namely, Afrasian-Lolo-Burmese as a negative example and Indo-European as a positive example. The question of what should be the most appropriate values that should make the test optimal is not addressed here. Ideally, to address this question, more data is needed with several positive and negative examples to search for optimal values of these parameters. Also, the exact values may require calibration according to the phylogenetic software used since there could be significant differences in the implementations. Secondly, while analyzing Nostratic languages, Uralic, an important language family, has not been included due to the selection criteria (\S\ref{subsec:data}) that the languages should have been attested before 10th century CE. To include Uralic, the (Nostratic) languages that are attested around the same period as the earliest attested ones from Uralic (roughly 1300 CE onwards) should be considered to make `fair' comparisons. 

\section*{Ethics Statement}
All the datasets are obtained from publicly available sources. Thus, there are no foreseen ethical considerations or conflicts of interest.
\bibliography{anthology,custom}

\begin{thebibliography}{50}
\expandafter\ifx\csname natexlab\endcsname\relax\def\natexlab#1{#1}\fi

\bibitem[{Akavarapu and
  Bhattacharya(2023)}]{akavarapu-bhattacharya-2023-cognate}
V.S.D.S.Mahesh Akavarapu and Arnab Bhattacharya. 2023.
\newblock \href {https://doi.org/10.18653/v1/2023.emnlp-main.423} {{Cognate
  Transformer for Automated Phonological Reconstruction and Cognate Reflex
  Prediction}}.
\newblock In \emph{Proceedings of the 2023 Conference on Empirical Methods in
  Natural Language Processing}, pages 6852--6862, Singapore. Association for
  Computational Linguistics.

\bibitem[{Akavarapu and
  Bhattacharya(2024)}]{akavarapu-bhattacharya-2024-automated}
V.S.D.S.Mahesh Akavarapu and Arnab Bhattacharya. 2024.
\newblock \href {https://aclanthology.org/2024.eacl-long.58} {{Automated
  Cognate Detection as a Supervised Link Prediction Task with Cognate
  Transformer}}.
\newblock In \emph{Proceedings of the 18th Conference of the European Chapter
  of the Association for Computational Linguistics (Volume 1: Long Papers)},
  pages 965--975, St. Julian{'}s, Malta. Association for Computational
  Linguistics.

\bibitem[{Anisimova and Gascuel(2006)}]{anisimova2006approximate}
Maria Anisimova and Olivier Gascuel. 2006.
\newblock {Approximate likelihood-ratio test for branches: A fast, accurate,
  and powerful alternative}.
\newblock \emph{Systematic Biology}, 55(4):539--552.

\bibitem[{Bishop and Friday(1987)}]{bishop1987tetrapod}
M~J Bishop and A~E Friday. 1987.
\newblock {Tetrapod relationships: The molecular evidence}.
\newblock \emph{Molecules and morphology in evolution: Conflict or compromise},
  pages 123--139.

\bibitem[{Bomhard and Kerns(1994)}]{bomhard1994nostratic}
Allan~R Bomhard and John~C Kerns. 1994.
\newblock \emph{{The Nostratic macrofamily: A study in distant linguistic
  relationship}}.
\newblock De Gruyter Mouton.

\bibitem[{Caldwell(1875)}]{caldwell1875comparative}
Robert Caldwell. 1875.
\newblock \emph{{A comparative grammar of the Dravidian or South-Indian family
  of languages}}.
\newblock Tr{\"u}bner.

\bibitem[{Campbell(1997)}]{campbell1997american}
Lyle Campbell. 1997.
\newblock \emph{{American Indian languages: The historical linguistics of
  Native America}}, volume~4.
\newblock Oxford University Press, USA.

\bibitem[{Campbell(2013)}]{campbell2013historical}
Lyle Campbell. 2013.
\newblock \emph{{Historical linguistics}}.
\newblock Edinburgh University Press.

\bibitem[{Felsenstein(1973)}]{felsenstein1973maximum}
Joseph Felsenstein. 1973.
\newblock {Maximum likelihood and minimum-steps methods for estimating
  evolutionary trees from data on discrete characters}.
\newblock \emph{Systematic Biology}, 22(3):240--249.

\bibitem[{Felsenstein(1981)}]{felsenstein1981evolutionary}
Joseph Felsenstein. 1981.
\newblock {Evolutionary trees from DNA sequences: A maximum likelihood
  approach}.
\newblock \emph{Journal of Molecular Evolution}, 17:368--376.

\bibitem[{Goldman et~al.(2000)Goldman, Anderson, and
  Rodrigo}]{goldman2000likelihood}
Nick Goldman, Jon~P Anderson, and Allen~G Rodrigo. 2000.
\newblock {Likelihood-based tests of topologies in phylogenetics}.
\newblock \emph{Systematic Biology}, 49(4):652--670.

\bibitem[{Greenberg(1963)}]{greenberg1963languages}
Joseph~H Greenberg. 1963.
\newblock {The languages of Africa}.
\newblock \emph{International Journal of American Linguistics}.

\bibitem[{Greenberg(1971)}]{greenberg1971indo}
Joseph~H Greenberg. 1971.
\newblock {The Indo-Pacific hypothesis}.
\newblock \emph{Current Trends in Linguistics}, 8:807--871.

\bibitem[{Greenberg(1987)}]{greenberg1987language}
Joseph~H Greenberg. 1987.
\newblock \emph{{Language in the Americas}}.
\newblock Stanford University Press.

\bibitem[{Greenberg(2000)}]{greenberg2000indo}
Joseph~H Greenberg. 2000.
\newblock \emph{{Indo-European and its closest relatives: The Eurasiatic
  language family, volume 1, grammar}}, volume~1.
\newblock Stanford University Press.

\bibitem[{Greenberg(2005)}]{greenberg2005genetic}
Joseph~H Greenberg. 2005.
\newblock \emph{{Genetic linguistics: Essays on theory and method}}.
\newblock OUP Oxford.

\bibitem[{Huelsenbeck and Bull(1996)}]{huelsenbeck1996likelihood}
John~P Huelsenbeck and JJ~Bull. 1996.
\newblock {A likelihood ratio test to detect conflicting phylogenetic signal}.
\newblock \emph{Systematic Biology}, 45(1):92--98.

\bibitem[{Huelsenbeck et~al.(1996)Huelsenbeck, Hillis, and
  Nielsen}]{huelsenbeck1996likelihood1}
John~P Huelsenbeck, David~M Hillis, and Rasmus Nielsen. 1996.
\newblock {A likelihood-ratio test of monophyly}.
\newblock \emph{Systematic Biology}, 45(4):546--558.

\bibitem[{J{\"a}ger(2015)}]{J_ger_2015}
Gerhard J{\"a}ger. 2015.
\newblock \href {https://doi.org/10.1073/pnas.1500331112} {{Support for
  linguistic macrofamilies from weighted sequence alignment}}.
\newblock \emph{Proceedings of the National Academy of Sciences},
  112(41):12752--12757.

\bibitem[{J{\"a}ger(2018)}]{J_ger_2018}
Gerhard J{\"a}ger. 2018.
\newblock \href {https://doi.org/10.1038/sdata.2018.189} {{Global-scale
  phylogenetic linguistic inference from lexical resources}}.
\newblock \emph{Scientific Data}, 5(1).

\bibitem[{J{\"a}ger(2019)}]{jager2019computational}
Gerhard J{\"a}ger. 2019.
\newblock {Computational Historical Linguistics}.
\newblock \emph{Theoretical Linguistics}, 45(3-4):151--182.

\bibitem[{J{\"a}ger(2022)}]{jager-2022-bayesian}
Gerhard J{\"a}ger. 2022.
\newblock \href {https://doi.org/10.18653/v1/2022.sigtyp-1.8} {{Bayesian
  Phylogenetic Cognate Prediction}}.
\newblock In \emph{Proceedings of the 4th Workshop on Research in Computational
  Linguistic Typology and Multilingual NLP}, pages 63--69, Seattle, Washington.
  Association for Computational Linguistics.

\bibitem[{Jukes et~al.(1969)Jukes, Cantor et~al.}]{jukes1969evolution}
Thomas~H Jukes, Charles~R Cantor, et~al. 1969.
\newblock {Evolution of protein molecules}.
\newblock \emph{Mammalian protein metabolism}, 3:21--132.

\bibitem[{Kassian et~al.(2015)Kassian, Zhivlov, and
  Starostin}]{kassian2015proto}
Alexei Kassian, Mikhail Zhivlov, and George Starostin. 2015.
\newblock {Proto-Indo-European-Uralic comparison from the probabilistic point
  of view}.
\newblock \emph{Journal of Indo-European Studies}, 43(3-4):301--347.

\bibitem[{Kessler(2001)}]{kessler2001significance}
Brett Kessler. 2001.
\newblock {The significance of word lists}.
\newblock \emph{Stanford}.

\bibitem[{Kessler(2007)}]{kessler-2007-word}
Brett Kessler. 2007.
\newblock \href {https://aclanthology.org/W07-1302} {{Word Similarity Metrics
  and Multilateral Comparison}}.
\newblock In \emph{Proceedings of Ninth Meeting of the {ACL} Special Interest
  Group in Computational Morphology and Phonology}, pages 6--14, Prague, Czech
  Republic. Association for Computational Linguistics.

\bibitem[{Kessler(2008)}]{kessler2008mathematical}
Brett Kessler. 2008.
\newblock {The Mathematical Assessment of Long-Range Linguistic Relationships}.
\newblock \emph{Language and Linguistics Compass}, 2(5):821--839.

\bibitem[{Kessler(2015)}]{kessler2015response}
Brett Kessler. 2015.
\newblock {Response to Kassian et al., Proto-Indo-European-Uralic comparison
  from the probabilistic point of view}.
\newblock \emph{Journal of Indo-European Studies}, 43(3-4):357--367.

\bibitem[{Kessler and Lehtonen(2006)}]{kessler2006multilateral}
Brett Kessler and Annukka Lehtonen. 2006.
\newblock {Multilateral comparison and significance testing of the Indo-Uralic
  question}.
\newblock \emph{Phylogenetic methods and the prehistory of languages}, pages
  33--42.

\bibitem[{Larkin et~al.(2007)Larkin, Blackshields, Brown, Chenna, McGettigan,
  McWilliam, Valentin, Wallace, Wilm, Lopez et~al.}]{larkin2007clustal}
Mark~A Larkin, Gordon Blackshields, Nigel~P Brown, R~Chenna, Paul~A McGettigan,
  Hamish McWilliam, Franck Valentin, Iain~M Wallace, Andreas Wilm, Rodrigo
  Lopez, et~al. 2007.
\newblock {Clustal W and Clustal X version 2.0}.
\newblock \emph{Bioinformatics}, 23(21):2947--2948.

\bibitem[{List(2010)}]{list2010sca}
Johann-Mattis List. 2010.
\newblock {SCA: Phonetic alignment based on sound classes}.
\newblock In \emph{European Summer School in Logic, Language and Information},
  pages 32--51. Springer.

\bibitem[{List(2012)}]{list-2012-lexstat}
Johann-Mattis List. 2012.
\newblock \href {https://aclanthology.org/W12-0216} {{LexStat: Automatic
  Detection of Cognates in Multilingual Wordlists}}.
\newblock In \emph{Proceedings of the {EACL} 2012 Joint Workshop of {LINGVIS}
  {\&} {UNCLH}}, pages 117--125, Avignon, France. Association for Computational
  Linguistics.

\bibitem[{List and Forkel(2021)}]{list2021lingpy}
Johann-Mattis List and Robert Forkel. 2021.
\newblock \href {https://lingpy.org} {{LingPy. A Python library for historical
  linguistics. Version 2.6.9}}.

\bibitem[{Ly-Trong et~al.(2022)Ly-Trong, Naser-Khdour, Lanfear, and
  Minh}]{ly2022alisim}
Nhan Ly-Trong, Suha Naser-Khdour, Robert Lanfear, and Bui~Quang Minh. 2022.
\newblock {AliSim: a fast and versatile phylogenetic sequence simulator for the
  genomic era}.
\newblock \emph{Molecular Biology and Evolution}, 39(5):msac092.

\bibitem[{Mailund and Pedersen(2004)}]{mailund2004qdist}
Thomas Mailund and Christian~NS Pedersen. 2004.
\newblock {QDist—Quartet distance between evolutionary trees}.
\newblock \emph{Bioinformatics}, 20(10):1636--1637.

\bibitem[{Nguyen et~al.(2015)Nguyen, Schmidt, Von~Haeseler, and
  Minh}]{nguyen2015iq}
Lam-Tung Nguyen, Heiko~A Schmidt, Arndt Von~Haeseler, and Bui~Quang Minh. 2015.
\newblock {IQ-TREE: a fast and effective stochastic algorithm for estimating
  maximum-likelihood phylogenies}.
\newblock \emph{Molecular Biology and Evolution}, 32(1):268--274.

\bibitem[{Oswalt(1970)}]{oswalt1970detection}
Robert~L Oswalt. 1970.
\newblock {The detection of remote linguistic relationships}.
\newblock \emph{Computer Studies in the Humanities and Verbal Behavior},
  3(3):117--129.

\bibitem[{Pompei et~al.(2011)Pompei, Loreto, and Tria}]{pompei2011accuracy}
Simone Pompei, Vittorio Loreto, and Francesca Tria. 2011.
\newblock {On the accuracy of language trees}.
\newblock \emph{PloS One}, 6(6):e20109.

\bibitem[{Poser and Campbell(2008)}]{poser2008language}
William Poser and Lyle Campbell. 2008.
\newblock {Language Classification: History and Methods}.

\bibitem[{Rama(2018)}]{rama-2018-similarity}
Taraka Rama. 2018.
\newblock \href {https://doi.org/10.18653/v1/K18-1027} {{Similarity Dependent
  Chinese Restaurant Process for Cognate Identification in Multilingual
  Wordlists}}.
\newblock In \emph{Proceedings of the 22nd Conference on Computational Natural
  Language Learning}, pages 271--281, Brussels, Belgium. Association for
  Computational Linguistics.

\bibitem[{Rama and List(2019)}]{rama-list-2019-automated}
Taraka Rama and Johann-Mattis List. 2019.
\newblock \href {https://doi.org/10.18653/v1/P19-1627} {{An Automated Framework
  for Fast Cognate Detection and {B}ayesian Phylogenetic Inference in
  Computational Historical Linguistics}}.
\newblock In \emph{Proceedings of the 57th Annual Meeting of the Association
  for Computational Linguistics}, pages 6225--6235, Florence, Italy.
  Association for Computational Linguistics.

\bibitem[{Rama et~al.(2018)Rama, List, Wahle, and
  J{\"a}ger}]{rama-etal-2018-automatic}
Taraka Rama, Johann-Mattis List, Johannes Wahle, and Gerhard J{\"a}ger. 2018.
\newblock \href {https://doi.org/10.18653/v1/N18-2063} {{Are Automatic Methods
  for Cognate Detection Good Enough for Phylogenetic Reconstruction in
  Historical Linguistics?}}
\newblock In \emph{Proceedings of the 2018 Conference of the North {A}merican
  Chapter of the Association for Computational Linguistics: Human Language
  Technologies, Volume 2 (Short Papers)}, pages 393--400, New Orleans,
  Louisiana. Association for Computational Linguistics.

\bibitem[{Ringe(1992)}]{ringe1992calculating}
Donald~A Ringe. 1992.
\newblock {On calculating the factor of chance in language comparison}.
\newblock \emph{Transactions of the American Philosophical Society},
  82(1):1--110.

\bibitem[{Ringe(1996)}]{ringe1996mathematics}
Donald~A Ringe. 1996.
\newblock {The mathematics of `Amerind'}.
\newblock \emph{Diachronica}, 13(1):135--154.

\bibitem[{Ringe and Eska(2013)}]{ringe2013historical}
Donald~A Ringe and Joseph~F Eska. 2013.
\newblock \emph{{Historical linguistics: Toward a twenty-first century
  reintegration}}.
\newblock Cambridge University Press.

\bibitem[{Sokal and Michener(1958)}]{Sokal1958ASM}
Robert~R. Sokal and Charles~Duncan Michener. 1958.
\newblock {A statistical method for evaluating systematic relationships}.
\newblock \emph{University of Kansas Science Bulletin}, 38:1409--1438.

\bibitem[{Tamura et~al.(2021)Tamura, Stecher, and Kumar}]{tamura2021mega11}
Koichiro Tamura, Glen Stecher, and Sudhir Kumar. 2021.
\newblock {MEGA11: molecular evolutionary genetics analysis version 11}.
\newblock \emph{Molecular Biology and Evolution}, 38(7):3022--3027.

\bibitem[{Turchin et~al.(2010)Turchin, Peiros, and
  Gell-Mann}]{turchin2010analyzing}
Peter Turchin, Ilia Peiros, and Murray Gell-Mann. 2010.
\newblock {Analyzing genetic connections between languages by matching
  consonant classes}.
\newblock \emph{Journal of Language Relationship}, (5 (48)):117--126.

\bibitem[{Wiley and Lieberman(2011)}]{wiley2011phylogenetics}
Edward~Orlando Wiley and Bruce~S Lieberman. 2011.
\newblock \emph{{Phylogenetics: Theory and practice of phylogenetic
  systematics}}.
\newblock John Wiley \& Sons.

\bibitem[{Wilks(1938)}]{Wilks_1938}
S.~S. Wilks. 1938.
\newblock \href {https://doi.org/10.1214/aoms/1177732360} {{The Large-Sample
  Distribution of the Likelihood Ratio for Testing Composite Hypotheses}}.
\newblock \emph{The Annals of Mathematical Statistics}, 9(1):60--62.

\end{thebibliography}

\appendix

\section{Analysis of Macro-Mayan and Amerind}
\label{sec:appendix}
\begin{figure}[h!]
    \centering
    \begin{subfigure}[t]{0.24\textwidth}
        \centering
        \includegraphics[width=\textwidth]{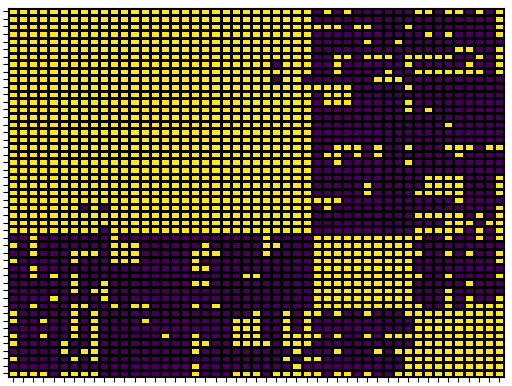}
        \caption{P1-Dolgo}
    \end{subfigure}%
    ~ 
   \begin{subfigure}[t]{0.24\textwidth}
        \centering
        \includegraphics[width=\textwidth]{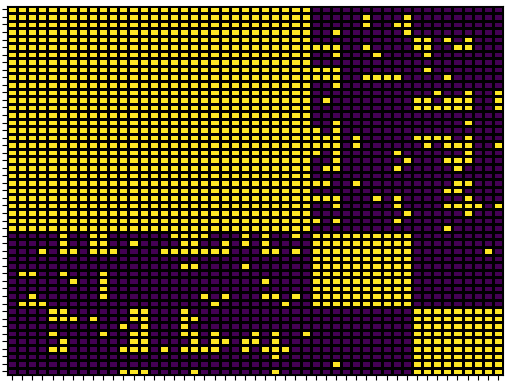}
        \caption{Turchin}
    \end{subfigure}%
    \\ 
    \begin{subfigure}[t]{0.24\textwidth}
        \centering
        \includegraphics[width=\textwidth]{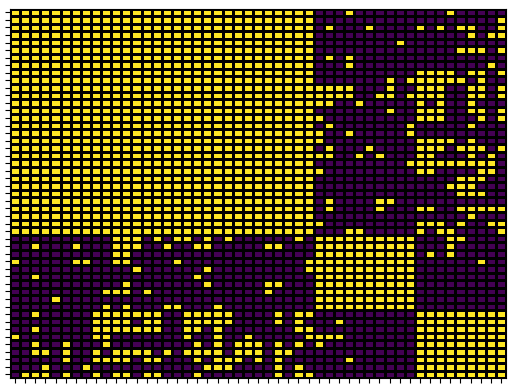}
        \caption{SCA}
    \end{subfigure}%
    ~ 
    \begin{subfigure}[t]{0.24\textwidth}
        \centering
        \includegraphics[width=\textwidth]{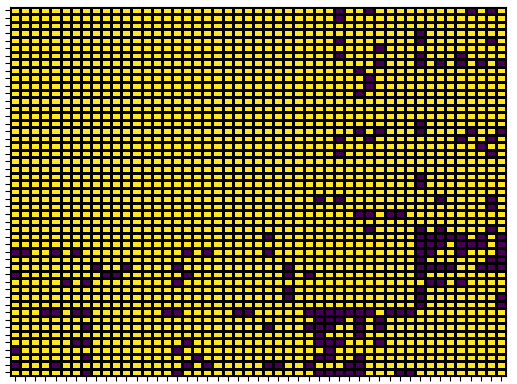}
        \caption{LexStat}
    \end{subfigure}%
    \caption{Bilateral (pairwise) significance among the languages of Macro-Mayan/Amerind grouping. The yellow shade implies that the relationship is statistically significant ($p < 0.05$), while the purple shade implies otherwise. While moving across the diagonal, the first cluster of significantly related languages is that of Mayan, the second is that of Mixe-Zoque and the thrid, Uto-Aztecan}
    \label{fig:mmu_biltrl}
\end{figure}

\begin{figure}[t]
    \centering
    \begin{subfigure}[t]{0.4\textwidth}
        \centering
        \includegraphics[width=\textwidth]{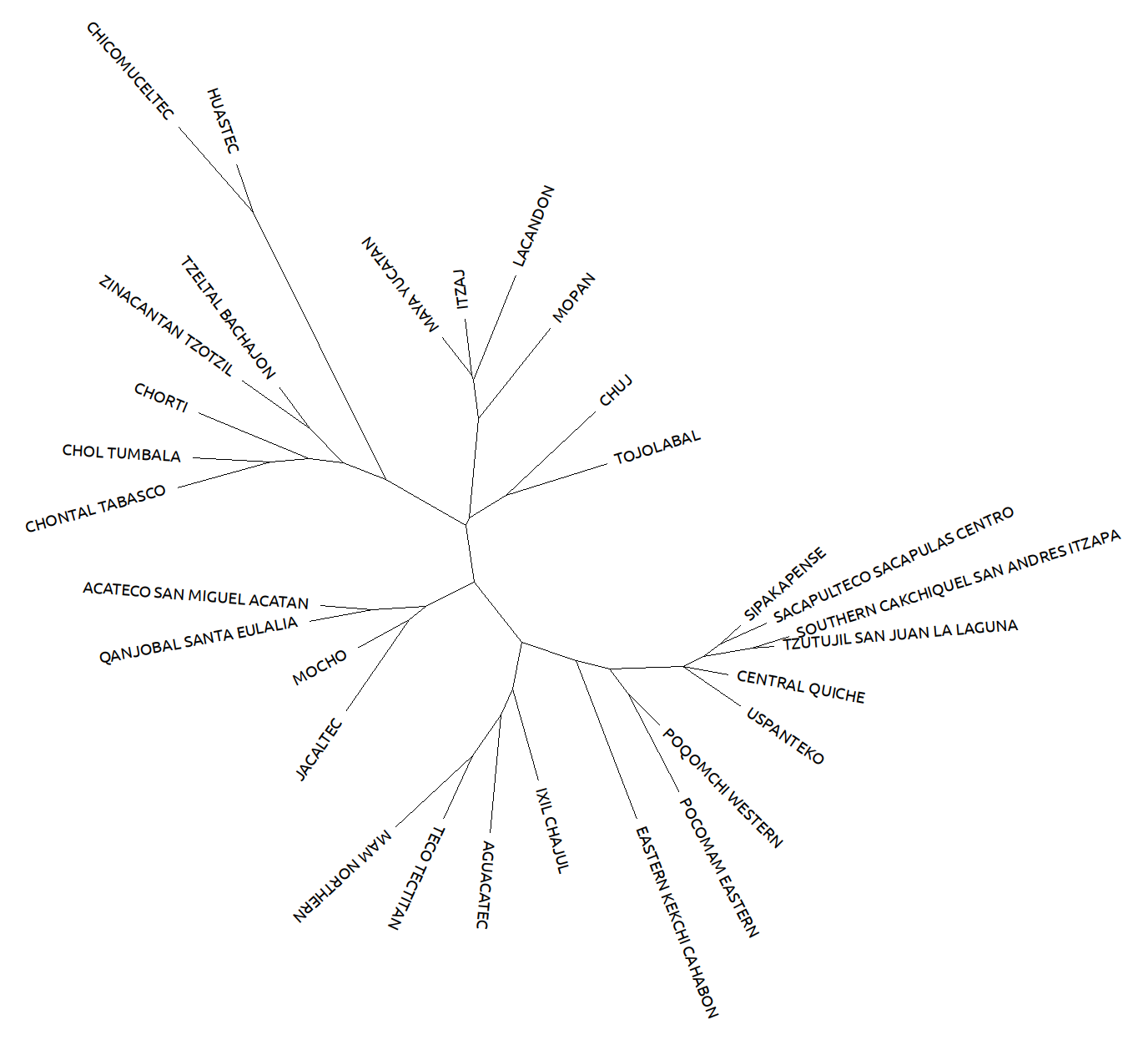}
        \caption{Mayan}
    \end{subfigure}%
    \\ 
    \begin{subfigure}[t]{0.4\textwidth}
        \centering
        \includegraphics[width=\textwidth]{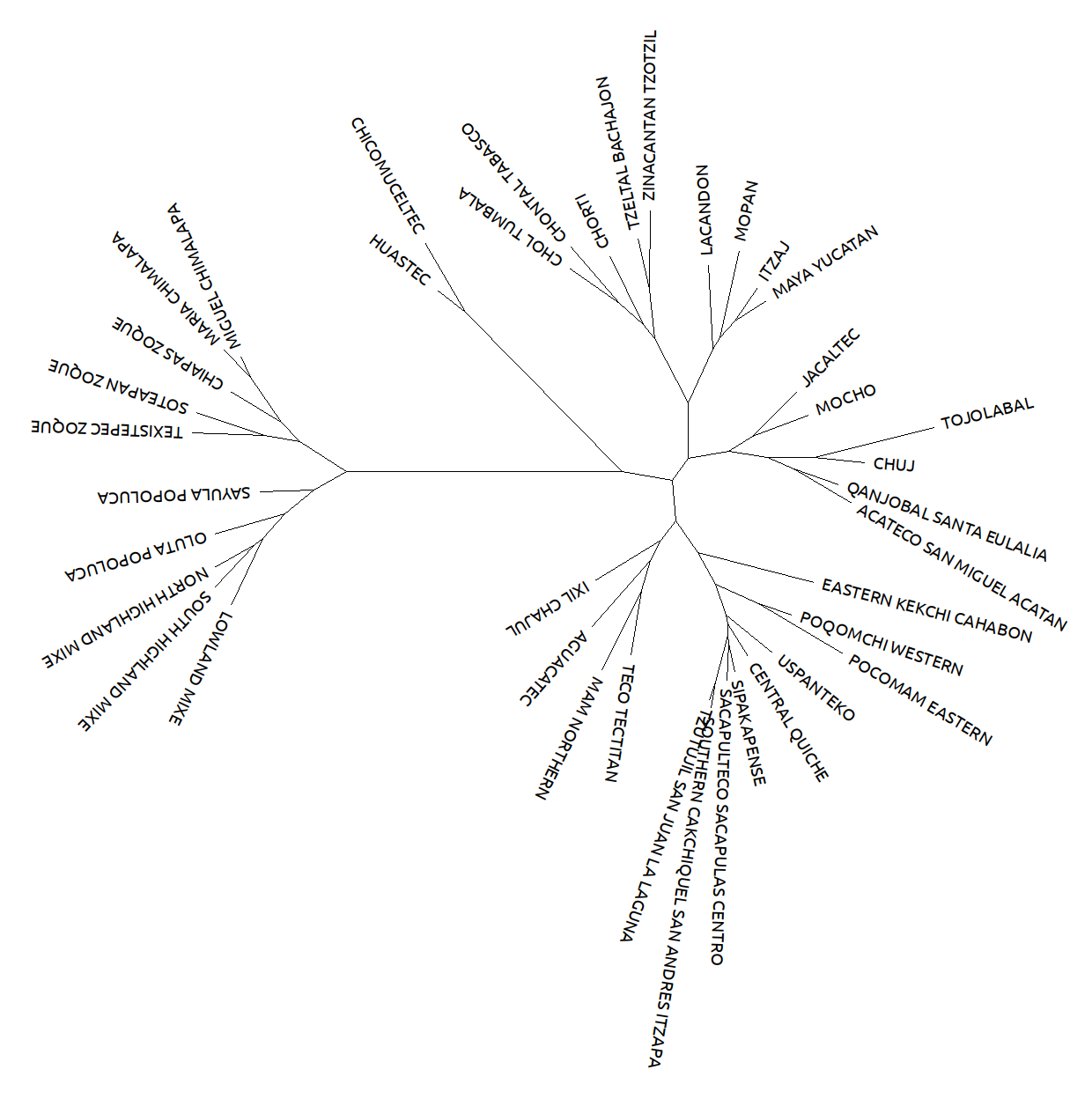}
        \caption{Mayan-Mixe-Zoque}
    \end{subfigure}
    \\ 
    \begin{subfigure}[t]{0.4\textwidth}
        \centering
        \includegraphics[width=\textwidth]{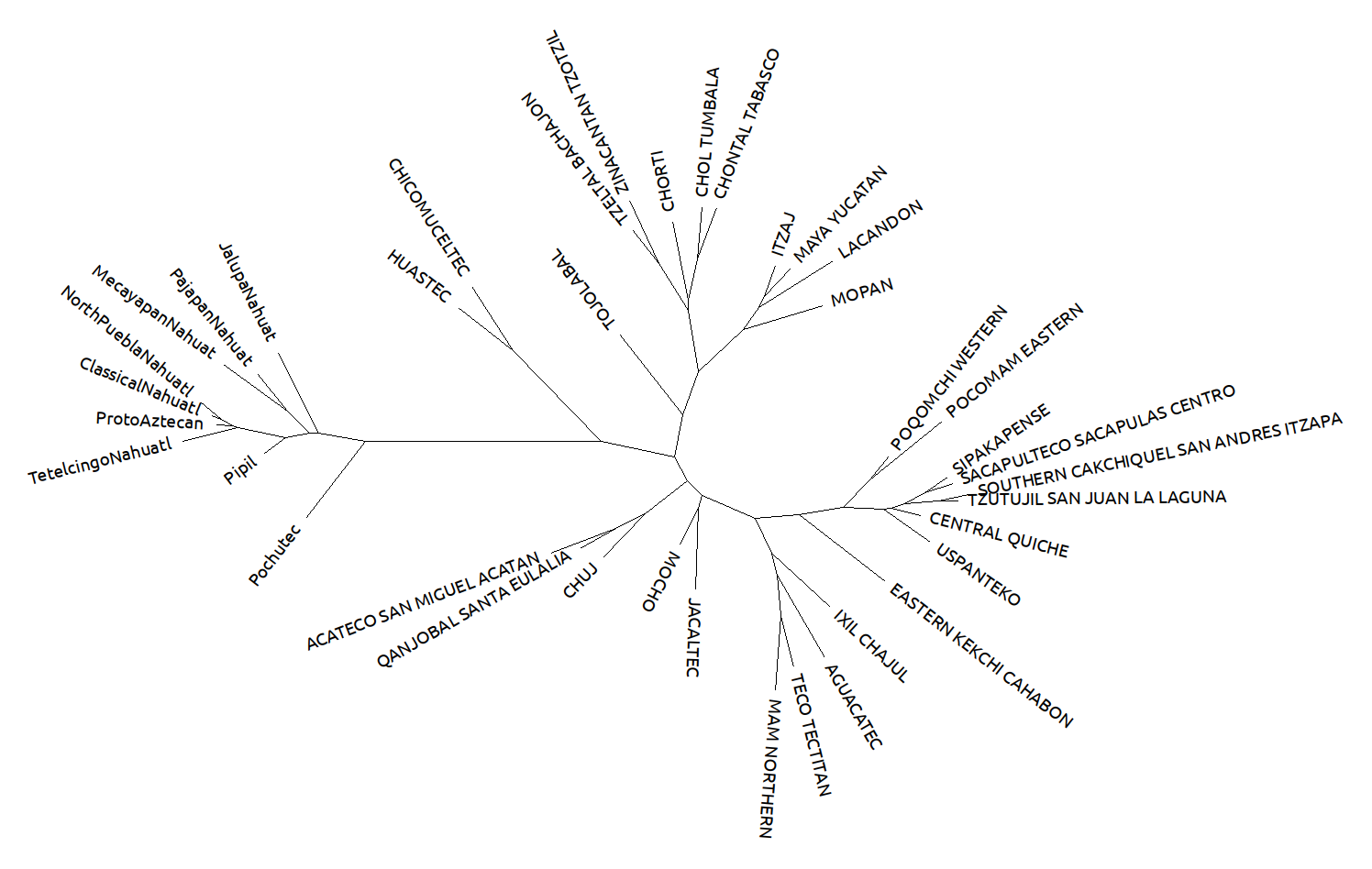}
        \caption{Mayan-Uto-Aztecan}
    \end{subfigure}
    \\ 
    \begin{subfigure}[t]{0.4\textwidth}
        \centering
        \includegraphics[width=\textwidth]{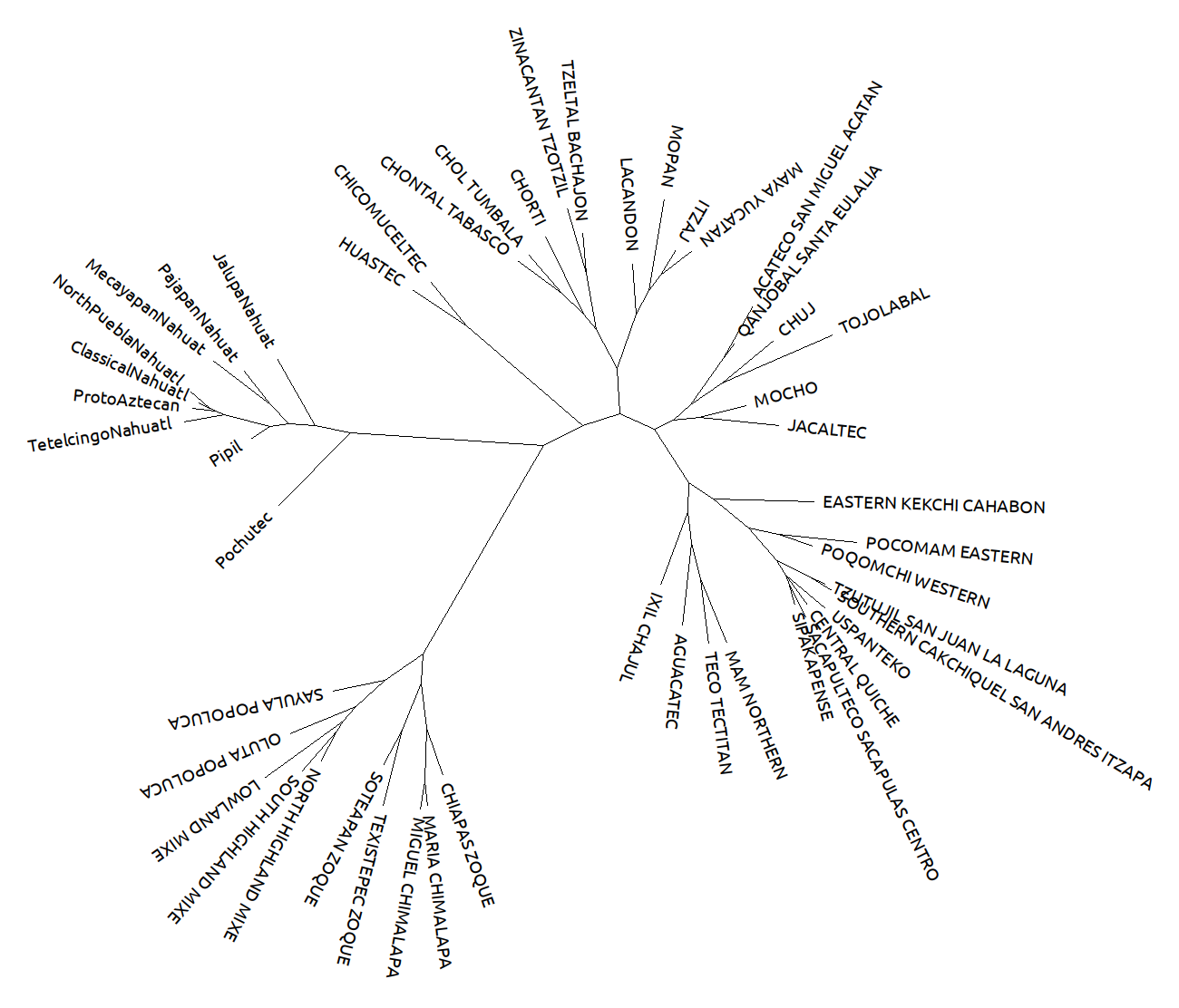}
        \caption{Mayan-Mixe-Zoque-Uto-Aztecan}
    \end{subfigure}
    \caption{Comparison of unrooted ML-trees on various groupings of Macro-Mayan/Amerind language families}
    \label{fig:mmu_mltree}
\end{figure}

\end{document}